# Linearity Properties of Bayes Nets with Binary Variables

**David Danks**
Institute for Human & Machine Cognition,
University of West Florida;
& University of California, San Diego

**Clark Glymour**
Carnegie-Mellon University;
Institute for Human & Machine Cognition,
University of West Florida;
& University of California, San Diego

## Abstract

It is "well known" that in linear models:

(1) testable constraints on the marginal distribution of observed variables distinguish certain cases in which an unobserved cause jointly influences several observed variables;

(2) the technique of "instrumental variables" sometimes permits an estimation of the influence of one variable on another even when the association between the variables may be confounded by unobserved common causes;

(3) the association (or conditional probability distribution of one variable given another) of two variables connected by a path or pair of paths with a single common vertex (a trek) can be computed directly from the parameter values associated with each edge in the trek;

(4) the association of two variables produced by multiple treks can be computed from the parameters associated with each trek; and

(5) the independence of two variables conditional on a third implies the corresponding independence of the sums of the variables over all units conditional on the sums over all units of each of the original conditioning variables.

These properties are exploited in search procedures. We show that (1) and (2) hold for all Bayes nets with binary variables. We further show that for Bayes nets parameterized as noisy-OR and noisy-AND gates, all of these properties save (4) hold.

## 1 INTRODUCTION

Linear models have special advantages for model search and for the estimation of causal effects, several of which are listed in the Abstract. Property (1) permits the detection of common causes via the Tetrad Representation Theorem, and in combination with properties (3) and (4) is sufficient for the determination of latent structural relations from rather weak background assumptions (Spirtes, *et al.*, 1993, 2001; Shafer, *et al.*., 1995). Property (2) provides a standard technique for estimating causal influence in econometrics, epidemiology and elsewhere. Property (5) is an essential assumption of many search methods that attempt to identify the causal structure of units from aggregated data. In particular, several proposed methods of discovering genetic regulatory networks from measurements of mRNA concentrations rely on such aggregated data.

In many models that are objects of automated search, for example networks for genetic regulation, it is assumed that the variables under study are binary. An important body of questions therefore concerns which of the properties of linear systems relevant to search hold for Bayes nets of binary variables, either in general or in an interesting class of special cases. Some results are known. For example the rules (3) and (4) for computing correlations in linear models are known to hold as well for singly trek-connected Bayes nets with binary variables, and counterexamples are known for networks that have multiple treks between pairs of variables (Pearl, 1988).

Techniques are known for using instrumental variables to bound causal effects in binary Bayes nets (Pearl, 2000). We supply a further result for Bayes nets of binary variables generally, and we discuss these properties for Bayes nets of binary variables parameterized as noisy-OR and noisy-AND gates, a parameterization of particular interest because of its use as a model of naïve human causal judgment (Cheng, 1997).

In what follows, all theorems and lemmas are given with (at most) proof sketches. A longer version of the paper, including full proofs, is available by contacting the first author.

## 2 GENERAL RESULTS

One technical notion and one Lemma will be used throughout this paper. A *trek* in a directed acyclic graph (DAG) is a directed path from one vertex to another, or a pair of directed paths terminating in two distinct vertices and intersecting in a single vertex. The unique vertex on any trek that has no edges (in the trek) directed into it is the *source* of the trek.

For example, in figure 2.1, $X \rightarrow Y \rightarrow Z$ and $W \leftarrow X \rightarrow Y$ are both treks (with $X$ as the source in both cases), but $W \rightarrow Z \leftarrow Y$ is not a trek.

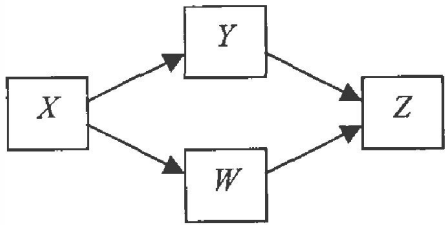

Figure 2.1: Sample graph

We also use the following general lemma, which connects conditional independence relations with correlation factorization in Bayes nets with only binary variables.

**Lemma** For any DAG with only binary variables, if $A \perp\!\!\!\perp C \mid B$, then $\rho(A, C) = \rho(A, B) * \rho(B, C)$.

### 2.1 A TETRAD REPRESENTATION THEOREM FOR BAYES NETS WITH BINARY VARIABLES

We can perform fast inference for unobserved common causes of the variables in a Bayes net when the Tetrad Representation Theorem (TRT) holds for that network (see Spirtes, *et al.*, 1993, 2001 for algorithmic details). The TRT is known to hold for linear systems; in this section we prove that it also holds for Bayes nets with only binary variables.

To formulate the TRT, we need the notion of a *choke point*: A variable $C$ is a choke point between two sets $\mathbf{I}$ and $\mathbf{J}$ if and only if for all variables $I \in \mathbf{I}$ and $J \in \mathbf{J}$, every trek between $I$ and $J$ includes $C$. Given this notion, we formulate the TRT for binary variables as follows:

**Tetrad Representation Theorem for Binary Variables**: In a DAG $G$, there is a choke point between two sets of variables, $\{I_1, I_2\}$ and $\{J_1, J_2\}$, if and only if $\rho_{11}\rho_{22} - \rho_{12}\rho_{21} = 0$ over a set of parameters of measure 1 (where $\rho_{ij}$ is the correlation between $I_i$ and $J_j$).

The two proofs of the TRT for linear systems have both used the generalized trek rule, which holds for all linear systems. The generalized trek rule states that:

$$\rho(X_0, X_n) = \sum_{t \in T} \prod_{i=1}^{t_n} \rho(X_{i-1}, X_i)$$

where $T$ is the set of all and only the treks connecting $X_0$ and $X_n$, and $t_n$ is the number of nodes on trek $t$. In fact, if the generalized trek rule holds for a system, then the TRT naturally follows, since the generalized trek rule is the only part of the TRT proof that depends on non-graphical properties. Unfortunately, the generalized trek rule does not hold in general for Bayes nets with binary variables (as we show below in section 4.1), and we therefore adopt a modified strategy.

Let $\mathbf{T}$ be the set of treks from $I$ to $J$, where $X_i$ ranges over the set of all variables on any $T \in \mathbf{T}$ (i.e., $X_i$ ranges over every variable, including $I$ and $J$, on all of the treks between $I$ and $J$). We then define the following two sets:

$\mathbf{U(T)} = \{<X_i, X_j> : \forall T \in \mathbf{T}(X_i \rightarrow X_j \in T)\}$

$\mathbf{S(T)} = \{<X_k, X_l> : [<X_k, X_l> \notin \mathbf{U(T)}] \& [\forall T \in \mathbf{T}(X_k, X_l \in T)] \& \neg \exists X_i[[\forall T \in \mathbf{T}(X_i \in T)] \& [X_i \text{ is between } X_k \text{ and } X_l]^1]\}$

These sets are actually quite easily described in English. $\mathbf{U(T)}$ consists of all of the pairs (i.e., directed edges) that appear in every trek from $I$ to $J$. $\mathbf{S(T)}$ consists of the first and last vertex of each portion of the treks that do not overlap. Note that at least one of the two sets will be non-empty (if $\mathbf{T}$ is non-empty). Figure 2.1.1 provides $\mathbf{U(T)}$ and $\mathbf{S(T)}$ for a sample graph.

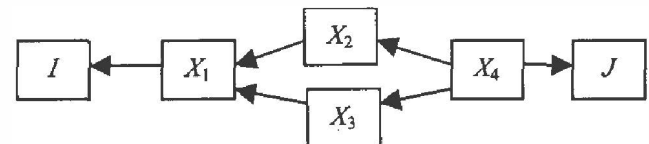

$\mathbf{U(T)} = \{<X_1, I>, <X_4, J>\}$

$\mathbf{S(T)} = \{<X_4, X_1>\}$

Figure 2.1.1: $\mathbf{U(T)}$ and $\mathbf{S(T)}$ for a sample graph

---

[1] Note that "between" is well-defined here, since $X_k$, $X_i$, and $X_l$ are on every trek, and each trek must go through them in the same order (see Shafer, *et al.*, 1995).

Note that we will omit the "(T)" when there is only one set of treks to consider. Given this notation, the following two theorems prove that a variant of the generalized trek rule holds for systems of binary variables.

**Theorem 2.1.1:**

Given the above notation, if **T** consists entirely of directed paths from $I$ to $J$,

$$\rho(I,J) = \left[\prod_{<X_i,X_j> \in \mathbf{U}} \rho(X_i, X_j)\right] * \left[\prod_{<X_i,X_j> \in \mathbf{S}} \rho(X_i, X_j)\right]$$

**Theorem 2.1.2:**

If **T** is the set of all treks between $I$ and $J$ (not necessarily all of which are directed paths), then

$$\rho(I,J) = \left[\prod_{<X_i,X_j> \in \mathbf{U}} \rho(X_i, X_j)\right] * \left[\prod_{<X_i,X_j> \in \mathbf{S}} \rho(X_i, X_j)\right]$$

**Proof sketch for Theorems 2.1.1 and 2.2.2:**

The above two theorems follow relatively directly from the earlier lemma. It turns out that, when we move along a trek between $I$ and $J$, we encounter the elements of **U** and **S** in the same order, regardless of which trek we choose. Furthermore, since any two elements of $\mathbf{U} \cup \mathbf{S} \cup \{I, J\}$ are d-separated by any element that falls between them, we can use the earlier lemma to factor the correlation between $I$ and $J$ into the above products.

Theorems 2.1.1 and 2.1.2 show that something similar to the generalized trek rule holds for systems of binary variables. It turns out that this variant is sufficient for the TRT, as the following two theorems show.

**Theorem 2.1.3:**

If there is at least one choke point between $\{I_1, I_2\}$ and $\{J_1, J_2\}$, then:

$\rho(I_1, J_1) * \rho(I_2, J_2) - \rho(I_1, J_2) * \rho(I_2, J_1) = 0$

**Theorem 2.1.4:**

If there is no choke point between $\{I_1, I_2\}$ and $\{J_1, J_2\}$, then for a measure 1 set of parameters,

$\rho(I_1, J_1) * \rho(I_2, J_2) - \rho(I_1, J_2) * \rho(I_2, J_1) \neq 0$

**Corollary 2.1.1:** (Tetrad Representation Theorem for binary variables)

There is at least one choke point between $\{I_1, I_2\}$ and $\{J_1, J_2\}$ iff:

$\rho(I_1, J_1) * \rho(I_2, J_2) - \rho(I_1, J_2) * \rho(I_2, J_1) = 0$

**Proof sketch for Theorems 2.1.3 and 2.1.4:**

First, we define $\mathbf{T}_{ij}$ to be the set of treks between $I_i$ and $J_j$. We can then use Theorems 2.1.1 and 2.1.2 to derive factorizations of the four correlations in the above theorems in terms of $\mathbf{U}(\mathbf{T}_{ij})$ and $\mathbf{S}(\mathbf{T}_{ij})$. Finally, note that the intersection of the $\mathbf{U}(\mathbf{T}_{ij})$ and $\mathbf{S}(\mathbf{T}_{ij})$ sets for all $i$ and $j$ will be non-empty if and only if there is at least one choke point. But if the intersection is non-empty, then the various factorizations of the correlations can be expressed using common terms which must satisfy the equality of theorem 2.1.3. If the intersection is empty (i.e., there is no choke point), then there will be terms that fail to appear in both correlation products, and those terms will only be equal for a measure 0 set of parameters.

## 2.2 AGGREGATION

In the case of gene expression research, our datapoints typically represent the sum of the variable values for a group of individuals. That is, we take data on several variables, but we actually receive only data summed or averaged over many individuals at once. Nevertheless, the aim of inquiry is a Bayes net representing the conditional independence and causal relations among the variables for *individual* units. So we pose the following question:

If $X \perp\!\!\!\perp Z \mid Y$ for each individual (in a large, i.i.d. sample of size $N$), is $\sum_{i=1}^{N} X_i \perp\!\!\!\perp \sum_{i=1}^{N} Z_i \mid \sum_{i=1}^{N} Y_i$, and conversely?

We abbreviate the conditional independence of the summed variables as $\Sigma X \perp\!\!\!\perp \Sigma Z \mid \Sigma Y$. We argue informally that for large $N$ almost certainly the conditional independence above holds for the summed variables if and only if it holds for the individual variables. For large $N$, the distribution of $\Sigma X$, $\Sigma Z$, $\Sigma Y$ is approximately normal by the Central Limit Theorem, and, to that approximation, a conditional independence holds if and only if the corresponding conditional covariance or partial correlation vanishes. We have the following formula for the conditional covariance of $\Sigma X$ and $\Sigma Z$:

$Cov(\Sigma X, \Sigma Z \mid \Sigma Y) = E(\Sigma X \& \Sigma Z \mid \Sigma Y) - E(\Sigma X \mid \Sigma Y)*E(\Sigma Z \mid \Sigma Y)$

The first term factors into: $E(\Sigma X \mid \Sigma Y) * E(\Sigma Z \mid \Sigma X \& \Sigma Y)$. Therefore, the covariation (and so also the correlation) is zero if and only if $E(\Sigma Z \mid \Sigma X \& \Sigma Y) = E(\Sigma Z \mid \Sigma Y)$.

We can express the expected value of $\Sigma Z$ as functions of the probabilities of $X$ and of $Y$ as:

$E(\Sigma Z) = N * P(Z = 1) =$    (2.2.1)

$N * [P(Z = 1 \mid X = 1, Y = 1) * P(Y = 1 \mid X = 1) * P(X = 1)$
$+ P(Z = 1 \mid X = 1, Y = 0) * P(Y = 0 \mid X = 1) * P(X = 1)$
$+ P(Z = 1 \mid X = 0, Y = 1) * P(Y = 1 \mid X = 0) * P(X = 0)$
$+ P(Z = 1 \mid X = 0, Y = 0) * P(Y = 0 \mid X = 0) * P(X = 0)]$.

and

$E(\Sigma Z) = N * P(Z = 1) =$    (2.2.2)

$N * [P(Z=1 \mid Y=1) * P(Y=1) + P(Z=1 \mid Y=0) * P(Y=0)]$.

Conditioning (2.2.2) on $\Sigma Y = N_Y$ results in

$E(\Sigma Z \mid \Sigma Y = N_Y) =$ (2.2.3)
$N * [P(Z=1 \mid Y=1) * N_Y/N + P(Z=1 \mid Y=0) * (1 - (N_Y/N))]$

Conditioning (2.2.1) on $\Sigma Y = N_Y$, $\Sigma X = N_X$ in the analogous way, rearranging and using the fact that $P(Z \mid X, Y) = P(Z \mid Y)$ also results in equation (2.2.3). Hence within the approximations noted, almost certainly $X \perp\!\!\!\perp Z \mid Y$ if and only if $\sum_{i=1}^{N} X_i \perp\!\!\!\perp \sum_{i=1}^{N} Z_i \mid \sum_{i=1}^{N} Y_i$.

## 3 BAYES NETS OF NOISY-OR/NOISY-AND GATES

Consider an arbitrary directed acyclic graph (DAG) whose vertices are binary variables taking values in $\{0,1\}$. We say a model is a noisy-OR and -AND gate model, or more briefly a Cheng model if, for each variable $X$, the set of parents of $X$, $Parents(X)$, can be partitioned into two sets, $GEN(X)$ and $PRE(X)$ such that:

$X = [U_x + \Sigma_{K \in GEN(X)} q_{KX} K] [\Pi_{L \in PRE(X)} (1 - q_{LX} L)]$

where all addition is Boolean addition, and $U_x$ is distributed independently of all variables other than $X$ and the descendants of $X$, and $q_{KX}$ and $q_{LX}$ are separate parameters for each variable $K$ and $L$, respectively, and all such parameters are jointly independent of each other and of all variables in the network. Intuitively, the variables in $GEN(X)$ and $U_X$ are generative or positive causes of $X$, while the variables in $PRE(X)$ prevent $X$ (taking $X = 1$ as the occurrence of $X$ or the marked case.) Again, intuitively, the probability that $q_{KX} = 1$ is the probability that, given that $K = 1$, $K$ causes $X = 1$, and the probability that $q_{LX} = 1$ is the probability that, given that $L = 1$, $L$ prevents $X = 1$ (Cheng, 1997). Sources of variation not represented in the network are required to be generative, since otherwise none of the parameters of the model can be estimated from observational data (Glymour, 1998). Such models have been applied in electrical engineering and developed as models of human judgment of non-interactive causal relations. Our concern is to find the linear analogies valid in such models.

### 3.1 INSTRUMENTAL VARIABLE CALCULATIONS

It has often been claimed that estimation of the edge parameters between two variables is impossible if there is an unobserved common cause of the two variables. This claim is false for situations in which we have an instrumental variable. Instrumental variable models have the graphical structure shown in figure 3.1.1.

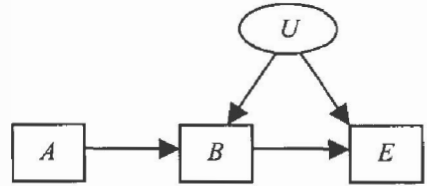

Figure 3.1.1: Instrumental variable graph

where $U$ is unobserved. The object is to estimate the conditional probability distribution of $E$ on values of $B$ determined by an intervention that randomizes $B$. Suppose all causes are generative, so that

$E = q_{be}B \oplus q_{ue}U$ and $B = q_{ab}A \oplus q_{ub}U$ (3.1.1)

where $\oplus$ is Boolean addition. Following Spirtes, et al. (1993, 2001), and Pearl (2000), we need to estimate:

$P_{B=0}(E=1) = P(q_{ue}U=1)$ and (3.1.2)
$P_{B=1}(E=1) = P(q_{be} \oplus q_{ue}U=1)$. (3.1.3)

It is easily verified that

$P(q_{ab}=1) =$ (3.1.4)
$[P(B=1 \mid A=1) - P(B=1 \mid A=0)] / [1 - P(B=1 \mid A=0)]$.

(The derivation is in Cheng, 1997). Substituting and factoring in (3.1.1), we have:

$E = q_{be} q_{ab}A \oplus (q_{be}q_{ub} \oplus q_{ue})U$ (3.1.5)

It follows by an analogous argument to that for (3.1.4) that

$P(q_{be}=1) * P(q_{ab}=1) = [P(E=1 \mid A=1) - P(E=1 \mid A=0)] / [1 - P(E=1 \mid A=0)]$

The ratio of (3.1.5) to (3.1.4) gives $P(q_{be} = 1)$. The r.h.s. of equation (3.1.2) is obtained by

$P(q_{ue}U=1) = P(q_{ue}U=1 \mid B=1) * P(B=1) + P(q_{ue}U=1 \mid B=0) * P(B=0)$

which after some algebra reduces to a formula in observed probabilities:

$P(q_{ue}U=1) = [P(E=1 \mid B=1) - P(q_{be}=1)] * P(B=1) / [1 - P(q_{be}=1)] + P(E=1 \mid B=0) * P(B=0)$

Hence the r.h.s. of (3.1.2) and (3.1.3) can be estimated. Analogous results are obtained with similar algebra when the influence of $B$ is preventive and $A$ is generative.

### 3.2 TREK RULES

In section 2.1, we noted the failure of the generalized trek rule for Bayes nets with only binary variables (demonstrated in section 4.1 below). Nevertheless, if we restrict our attention to these Cheng models, then we can derive a closed-form expression for the correlation of two variables connected by a single trek. We assume the following typical structure:

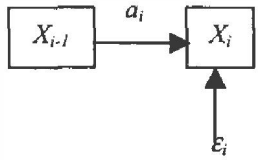

Figure 3.2.1: Typical Cheng model unit

where the response functions (and associated probabilities) are:

*Noisy-OR gate:*

$X_i = a_i X_{i-1} \oplus \varepsilon_i$

$P(X_i) = P(a_i) * P(X_{i-1}) + P(\varepsilon_i) - P(a_i) * P(X_{i-1}) * P(\varepsilon_i)$

*Noisy-AND gate:*

$X_i = \varepsilon_i \bullet (1 - a_i X_{i-1})$

$P(X_i) = P(\varepsilon_i) * [1 - P(a_i) * P(X_{i-1})]$

**Theorem 3.2.1:**

If a directed path of length $n \geq 1$ composed of noisy-OR and noisy-AND gates (in any combination and order) is the only trek between $X_0$ and $X_n$, then:

$$\rho(X_0, X_n) = \left[\prod_{i=1}^{n} P(a_i) * g(i)\right] * \frac{\sqrt{P(X_0) * [1 - P(X_0)]}}{\sqrt{P(X_n) * [1 - P(X_n)]}},$$

where $g(i) = \begin{cases} [1 - P(\varepsilon_i)], & \text{if the } i\text{-th gate is noisy-OR; or} \\ -P(\varepsilon_i), & \text{if the } i\text{-th gate is noisy-AND.} \end{cases}$

This closed-form expression is only for the correlation of the ends of a directed path, and not for two singly trek-connected variables. The earlier lemma, as well as results from Pearl (1988), enables us to derive the following two factorization theorems.

**Theorem 3.2.2:**

If a directed path of length $n \geq 1$ composed of noisy-OR and noisy-AND gates (in any combination and order) is the only trek between $X_0$ and $X_n$, then:

$$\rho(X_0, X_n) = \prod_{i=1}^{n} \rho(X_{i-1}, X_i)$$

**Theorem 3.2.3:**

If a trek of length $n \geq 1$ composed of noisy-OR and noisy-AND gates (in any combination and order) with $X_k$ as the source of the trek ($n \geq k \geq 0$) is the only trek between $X_0$ and $X_n$, then:

$$\rho(X_0, X_n) = \prod_{i=1}^{n} \rho(X_{i-1}, X_i)$$

We can then use these factorizations to derive a closed-form expression for two singly trek-connected variables. That formula is given by the following corollary:

**Corollary 3.2.1:** (follows directly from theorems 3.2.3 and 3.2.1)

If a trek of length $n \geq 1$ composed of noisy-OR and noisy-AND gates (in any combination and order) with source $X_k$ ($n \geq k \geq 0$) is the only trek between $X_0$ and $X_n$, then:

$$\rho(X_0, X_n) = \left[\prod_{i=0}^{k-1} P(a_i) * g(i)\right] * \left[\prod_{i=k+1}^{n} P(a_i) * g(i)\right] * \frac{P(X_k) * [1 - P(X_k)]}{\sqrt{P(X_0) * [1 - P(X_0)]} * \sqrt{P(X_n) * [1 - P(X_n)]}}$$

where $g(i) = \begin{cases} [1 - P(\varepsilon_i)], & \text{if the } i\text{-th gate is noisy-OR; or} \\ -P(\varepsilon_i), & \text{if the } i\text{-th gate is noisy-AND.} \end{cases}$

## 4 COUNTEREXAMPLES

We might naturally wonder whether the results provided in this paper are all "as good as it gets." For example, perhaps the closed-form expression given in Corollary 3.2.1 also applies to multiply trek-connected variables. This section provides counter-examples demonstrating the limits of the above results. The trek rules for singly connected Cheng models do not generalize. Further, Cheng models make it easy to show that the aggregation invariance that holds in all Bayes nets with binary variables when conditioning on a single variable does not hold when conditioning on multiple variables.

### 4.1 FAILURE OF THE TREK RULE

The above trek rule (for singly trek-connected variables) does not generalize to multiply trek-connected variables in noisy-AND/OR networks. That is: If $T$ is the set of all and only the treks between $X_0$ and $X_n$, and $|T| > 1$, then it is not necessarily the case that: $\rho(X_0, X_n) = \sum_{t \in T} \rho_t(X_0, X_n)$ (where $\rho_t(X_0, X_n)$ is the correlation between $X_0$ and $X_n$ if trek $t$ were the only trek).

Consider the following graph composed solely of noisy-AND gates (the $a_i$ and $\varepsilon_i$ terms are left out for simplicity):

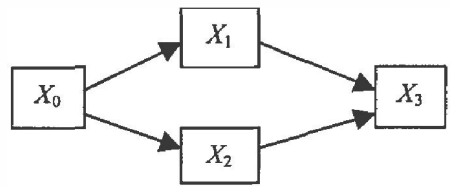

Figure 4.1.1: Counterexample to trek rule

So, the equations for the dependent variables are:

$X_1 = [1 - a_1 X_0] \bullet \varepsilon_1$

$X_2 = [1 - a_2X_0] \bullet \varepsilon_2$

$X_3 = [1 - a_{31}X_1] \bullet [1 - a_{32}X_2] \bullet \varepsilon_3$

We only need to determine $\rho(X_0, X_3)$ directly, since we can use theorem 3.2.1 to compute the correlations along each trek (since each is a directed path). When we substitute the equations for $X_1$ and $X_2$ into the equation for $X_3$, we get:

$P(X_3) = [1 - P(a_{31})*P(\varepsilon_1) * [1 - P(a_1) * P(X_0)]] * [1 - P(a_{32}) * P(\varepsilon_2) * [1 - P(a_2) * P(X_0)]] * P(\varepsilon_3)$

After much algebra, we can derive the following formula for the total correlation:

$$\rho(X_0, X_3) = P(\varepsilon_3) * Q * \frac{\sqrt{P(X_0)*[1-P(X_0)]}}{\sqrt{P(X_3)*[1-P(X_3)]}},$$

where $Q = [P(\varepsilon_1)*P(a_1)*P(a_{31}) + P(\varepsilon_2)*P(a_2)*P(a_{32}) - P(\varepsilon_1)*P(\varepsilon_2)*P(a_1)*P(a_{31})*P(a_{32}) - P(\varepsilon_1)*P(\varepsilon_2)*P(a_2)*P(a_{31})*P(a_{32}) + P(\varepsilon_1)*P(\varepsilon_2)*P(a_1)*P(a_2)*P(a_{31})*P(a_{32})*P(X_0)]$

Using Theorem 3.2.1 to compute the correlations along each individual trek, we have

$$\rho_1(X_0, X_3) + \rho_2(X_0, X_3) = P(\varepsilon_3) * W * \frac{\sqrt{P(X_0)*[1-P(X_0)]}}{\sqrt{P(X_3)*[1-P(X_3)]}},$$

where $W = [P(\varepsilon_1)*P(a_1)*P(a_{31}) + P(\varepsilon_2)*P(a_2)*P(a_{32})]$.

Therefore, we can see that the generalized trek rule will hold for this case if and only if $Q = W$, which (since we assume non-extremal probabilities) is true if and only if:

$[P(a_1) + P(a_2) - P(a_1)*P(a_2)*P(X_0)] = 0$

This equality cannot possibly be satisfied (since $P(a_1)*P(a_2)*P(X_0) < P(a_1)$ and $P(a_1)*P(a_2)*P(X_0) < P(a_2)$). Therefore, the generalized trek rule does not hold for all graphs composed of noisy-OR and noisy-AND gates.

### 4.2 FAILURE OF AGGREGATION

As with the trek rule, the results on aggregation do not generalize to graphs with multiply trek-connected variables. That is: If $X$ is an ancestor of $Z$, and $Y_1, \ldots, Y_n$ ($n > 1$) are the parents of $Z$, then it is not necessarily the case that $\rho(\Sigma X, \Sigma Z \mid \Sigma Y_1, \ldots, \Sigma Y_n) = 0$. We prove this fact through counterexample. Consider the following graph:

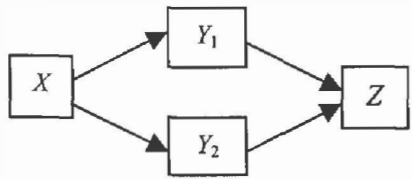

Figure 4.2.1: Counterexample to aggregation

We have the following formula:

$Cov(\Sigma X, \Sigma Z \mid \Sigma Y_1, \Sigma Y_2) = E(\Sigma X \& \Sigma Z \mid \Sigma Y_1, \Sigma Y_2) - E(\Sigma X \mid \Sigma Y_1, \Sigma Y_2) * E(\Sigma Z \mid \Sigma Y_1, \Sigma Y_2)$

The first term in the formula factors into: $E(\Sigma X \mid \Sigma Y_1, \Sigma Y_2) * E(\Sigma Z \mid \Sigma X, \Sigma Y_1, \Sigma Y_2)$. Therefore, the covariance (and hence the correlation) equals zero just in case:

$E(\Sigma Z \mid \Sigma X, \Sigma Y_1, \Sigma Y_2) = E(\Sigma Z \mid \Sigma Y_1, \Sigma Y_2)$.

Now consider the left-hand side of the equation. If we assume that there are $N$ individuals in the summation, that the summations are given by $N_X$, $N_{Y1}$, and $N_{Y2}$, and that all of the connections are noisy-AND gates, then we have:

$E(\Sigma Z \mid \Sigma X, \Sigma Y_1, \Sigma Y_2) = N * [(1 - P(a_{Y1}) * P(Y_1)) * (1 - P(a_{Y2})*P(Y_2)) * P(\varepsilon_Z)] = P(\varepsilon_Z) * [N - P(a_{Y1})*N_{Y1} - P(a_{Y2})*N_{Y2} - P(a_{Y1})*P(a_{Y1})*N*P(Y_1 \& Y_2)]$.

Now, $P(Y_1 \& Y_2)$ is a function of $X$, and so we can reduce $E(\Sigma Z \mid \Sigma X, \Sigma Y_1, \Sigma Y_2)$ to a formula having only known values (including $N_X$, $N_{Y1}$, and $N_{Y2}$).

Consider a similar operation on $E(\Sigma Z \mid \Sigma Y_1, \Sigma Y_2)$. In this case, our simplification must stop with a $P(Y_1 \& Y_2)$ term still in the formula. That is, we cannot determine whether, in fact, these two equations are equal. It depends on the probability of the joint occurrence of $Y_1$ and $Y_2$, which we do not know.

## 5 COMMENTS

The counterexample to aggregation invariance argues that, except in special cases, attempts to infer an underlying structure among binary variables from aggregated data ought to be suspect. On the positive side, the explicit characterization of trek rules and the applicability of instrumental variables to noisy-OR/noisy-AND gate models may be of use both in the design of psychological experiments and in data analysis where such parameterizations are plausible.

The most important positive result in this paper is surely the extension of the Tetrad Representation Theorem to systems of binary variables. Combined with the absence of conditional independence relations among the measured variables (as in Spirtes, et al., 1993, 2001) it provides a necessary and sufficient condition (assuming "faithfulness" – see Spirtes, et al., 1993, 2001) for four measured variables in a structure of binary variables to have a single unmeasured common cause. The applicability of the result bears comparison with recent statistical work (Junker and Ellis, 1997) that provides a sufficient condition (implicitly with the same faithfulness assumption) for a single common cause given an infinite sequence of measured variables. An interesting open question concerns whether results similar to the TRT can be obtained for models now popular in psychometrics in which the probability distribution on measured binary variables is a function of a continuous latent variable.